%% file: main.tex
\documentclass[10pt,twocolumn,letterpaper]{article}

\usepackage[pagenumbers]{cvpr}      
\input{preamble}

\definecolor{cvprblue}{rgb}{0.21,0.49,0.74}
\usepackage[pagebackref,breaklinks,colorlinks,allcolors=cvprblue]{hyperref}

\def\confName{CVPR}
\def\confYear{2027}

\title{Reading Uncertainty from Frozen World Models for Visual Planning}

\author{Shicheng Yin$^{1*}$\quad Hao Li$^{1*}$\quad Kaixuan Yin$^{1}$\quad Yang Liu$^{1,2\dagger}$\\
$^{1}$Sun Yat-sen University\quad $^{2}$X-Era AI Lab\\
{\small \{yinshch5, lihao323, yinkx3\}@mail2.sysu.edu.cn,\ \ liuy856@mail.sysu.edu.cn}\\
{\small Code: \href{https://github.com/HCPLab-SYSU/RU-WM}{\textcolor{black}{github.com/HCPLab-SYSU/RU-WM}}}}

\makeatletter\@namedef{ver@eso-pic.sty}{2020/10/14}\makeatother
\usepackage{pdfpages}
\begin{document}
\maketitle
\input{figs/teaser_lift}
{\let\thefootnote\relax\footnotetext{$^{*}$Equal contribution.\quad $^{\dagger}$Corresponding author.}}
\input{sec/0_abstract}
\input{sec/1_intro}
\input{sec/2_related}
\input{sec/3_method}
\input{sec/4_experiments}
\input{sec/5_conclusion}
{
    \small
    \bibliographystyle{ieeenat_fullname}
    \bibliography{main}
}
\clearpage
\includepdf[pages=-]{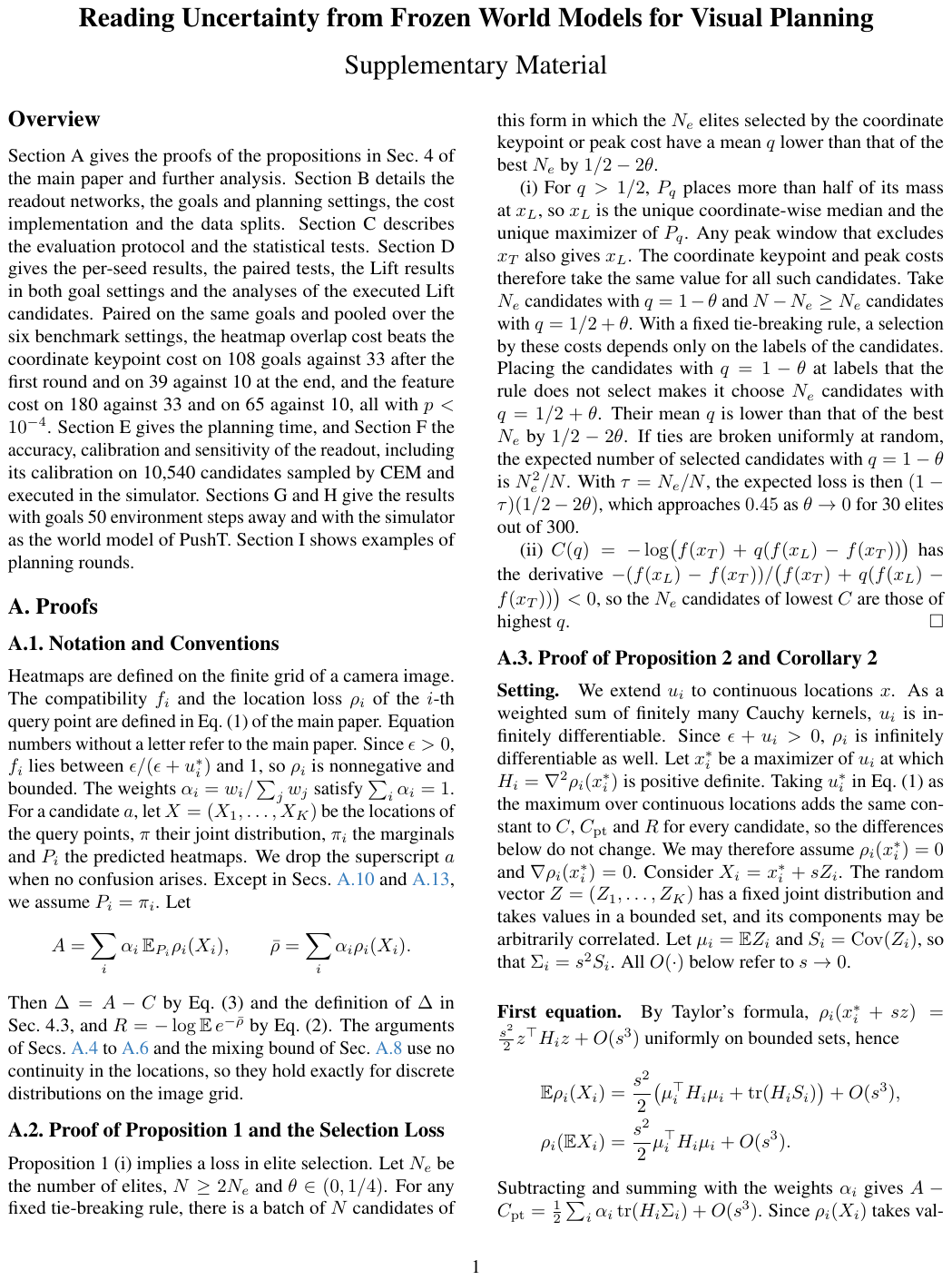}
\end{document}

%% file: preamble.tex
\usepackage{cuted}    \usepackage{currfile} \usepackage{caption}
\usepackage{multirow}
\usepackage{textcomp,amsmath,amssymb,booktabs,array,tabularx,makecell}

\usepackage{amsthm}

\newtheorem{corollary}{Corollary}
\newtheorem{proposition}{Proposition}

\DeclareMathOperator{\MSE}{MSE}

\input{figures_preamble}

\usepackage[protrusion=false]{microtype}

%% file: figures_preamble.tex
\usepackage{tikz}
\usetikzlibrary{positioning,arrows.meta,fit,calc,backgrounds,decorations.pathreplacing,shapes.geometric}

\definecolor{bpOurs}{RGB}{0,114,178}      \definecolor{bpPrior}{RGB}{213,94,0}      \definecolor{bpData}{RGB}{0,158,115}      \definecolor{bpAccent}{RGB}{230,159,0}    \definecolor{bpGray}{RGB}{110,110,110}    \definecolor{bpLight}{RGB}{238,238,238}   \colorlet{bpOursFill}{bpOurs!18}
\colorlet{bpPriorFill}{bpPrior!15}
\colorlet{bpDataFill}{bpData!15}

\tikzset{
  bp/base/.style   ={draw=black!70, line width=0.5pt, rounded corners=2pt, align=center,
                     font=\sffamily\scriptsize, minimum height=0.6cm, inner sep=3pt},
  bp/module/.style ={bp/base, fill=white},
  bp/frozen/.style ={bp/base, fill=bpLight, draw=bpGray, text=black!80},
  bp/ours/.style   ={bp/base, fill=bpOursFill, draw=bpOurs, line width=0.7pt},
  bp/prior/.style  ={bp/base, fill=bpPriorFill, draw=bpPrior},
  bp/data/.style   ={bp/base, fill=bpDataFill, draw=bpData},
  bp/flow/.style   ={-{Latex[length=1.6mm]}, line width=0.6pt, black!75},
  bp/note/.style   ={font=\sffamily\tiny, text=bpGray, align=center},
  bp/panel/.style  ={font=\sffamily\footnotesize\bfseries, anchor=west},
  bp/frame/.style  ={draw=black!70, fill=white, line width=0.4pt},
}

%% file: figs/teaser_lift.tex
\begin{strip}
\centering
\includegraphics[width=\textwidth,trim=0 5 0 10.5,clip]{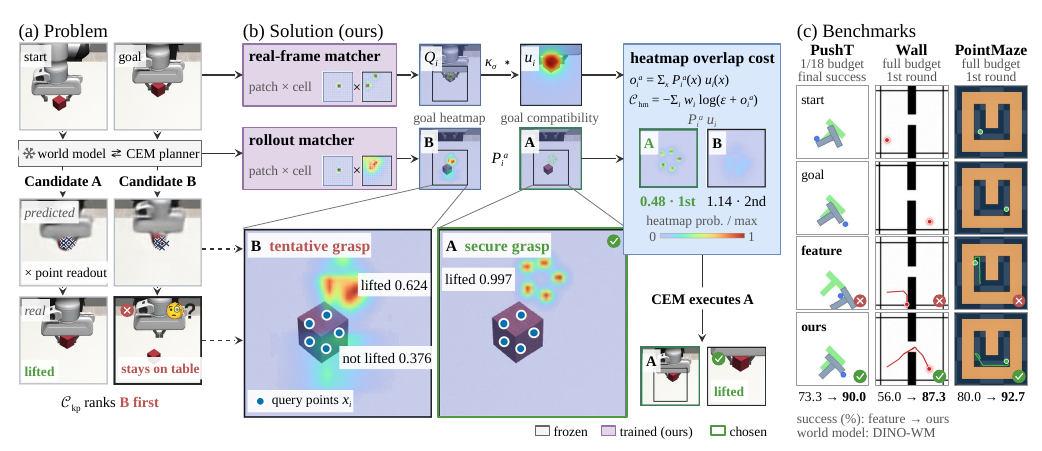}
\captionsetup{hypcap=false}
\captionof{figure}{\textbf{A point readout can rank a failing grasp first. The heatmap overlap cost ranks grasps by their chance of success.}
(a) Candidates A and B come from the same start and have nearly the same coordinate readouts.
The coordinate keypoint cost $\mathcal{C}_{\mathrm{kp}}$ ranks B first, and B leaves the block on the table.
(b) The rollout matcher reads their predictions as heatmaps $P_i^a$, which put 0.997 of the probability of A and 0.624 of B where the block is lifted.
The heatmap overlap cost compares them with the goal compatibility $u_i$ and ranks A first. CEM executes A.
(c) Benchmark examples where the feature cost fails and ours succeeds, with success rates (\%) on DINO-WM (\cref{tab:main,tab:budget}).}
\captionsetup{hypcap=true}
\label{fig:teaser}
\end{strip}

%% file: sec/0_abstract.tex
\begin{abstract}
A frozen visual world model makes one deterministic prediction for each candidate action, and this prediction is imperfect.
Two candidates can therefore look equally successful in the prediction, but one may fail.
A cost that scores only the predicted outcome cannot tell them apart.
For example, a coordinate keypoint cost can rank a grasp that leaves a block on the table above one that lifts it.
We train a readout network on rollouts of the frozen model.
Given a prediction, it outputs a heatmap of where each query point will actually be.
We rank candidates by the overlap between predicted and goal heatmaps.
This gives CEM a more discriminative optimization signal where point costs are flat.
Neither the world model nor the planner is changed.
Our analysis shows that a point readout loses two kinds of information.
It ties candidates that differ only in their probability of success, and near the goal it ignores how dispersed the prediction is.
The heatmap overlap cost keeps both.
In a block-lifting task, it succeeds in 86.3\% of episodes, compared with 59.8\% for the coordinate keypoint cost and 73.5\% when the same heatmaps are scored by their peaks.
It also has the highest average success rate on PushT, Wall and PointMaze with DINO-WM and DDP-WM.
\end{abstract}

%% file: sec/1_intro.tex
\section{Introduction}
\label{sec:intro}

Visual world models predict the effect of actions in the feature space of a pretrained visual encoder~\cite{dinowm,ddpwm,vjepa2}.
A planner such as CEM samples candidate action sequences and ranks their predictions against a goal image with a cost.
The world model is frozen, and its prediction is deterministic but imperfect.
Given the prediction, the true outcome of a candidate is therefore uncertain.
This uncertainty comes from the error of the world model.
A candidate that looks successful in the prediction may still fail.
CEM follows the top-ranked candidates, so a cost that ignores this uncertainty can lead it to an unreliable action.
We ask whether a cost can keep how likely each outcome is.

Existing costs score only the predicted outcome.
In \cref{fig:teaser}, candidates A and B have nearly the same coordinate readouts, and the coordinate keypoint cost ranks B first.
But the heatmaps of B put 0.38 of the probability where the block is not lifted, and B fails.
The most common cost is the distance between predicted and goal features~\cite{dinowm,ddpwm,vjepa2,terver2026jepawms,zhang2026hwm}.
It also counts changes in the background and in irrelevant objects, which do not advance the task.

We instead read out where the task-relevant parts will actually be.
Our probabilistic keypoint interface takes any query point in the image and returns a heatmap of its future location (\cref{fig:pipeline}).
Earlier keypoint methods infer how points move from rigid-body assumptions~\cite{kpam,rekep} or build the points into the training of the dynamics model~\cite{kitf,dexwm}.
Our interface needs neither and reads the predictions of a frozen world model.
Its rollout matcher is trained on rollouts of the world model itself.
The labels are point tracks obtained with SAM~3~\cite{sam3} and CoTracker3~\cite{cotracker3}, so no simulator state is needed.
Under the training action distribution, the optimum of its cross-entropy loss is the distribution of the true location given the prediction.
On executed candidates, its probabilities are calibrated (\cref{fig:lift-prob}a).

We score a candidate by the overlap between the predicted heatmap of each query point and its heatmap in the goal image.
The overlap is the expected compatibility of the predicted location with the goal.
It uses the entire heatmap, so it separates candidates that look equally successful but differ in their chance of success.
A grasp that may fail then ranks below a secure grasp.

We then ask what a planner loses when each prediction is reduced to a point.
When a part may end up in one of two distant places, a point readout keeps only the more likely one.
Candidates with different probabilities of success then get the same cost.
The heatmap overlap cost ranks them by this probability (\cref{prop:gap}).
Near the goal, a point readout ignores how spread out the prediction is.
Of two candidates predicted at the same location, it cannot prefer the more certain one (\cref{thm:local}).
The task needs all parts in place together, but each heatmap describes one part.
Scoring each part by its own heatmap gives the same order as the joint goal whenever two costs differ by more than a gap given by the heatmaps (\cref{cor:select}).
We measure each of these effects on real predictions.

We evaluate on lifting a block, the contact-rich PushT and two navigation tasks, with the world models of DINO-WM and DDP-WM.
On PushT, we also replace the world model by the exact simulator as a control.
On executed candidates, the probability of lifting read from the heatmaps predicts the outcome better than the coordinate readout at every start.
On PushT with goals 25 steps away, the heatmap overlap cost with 1/18 of the computation exceeds every baseline at full budget.

Our contributions are as follows.
\begin{itemize}
\item A probabilistic keypoint interface that turns the prediction of a frozen world model into the distribution of the future location of any query point.
\item The heatmap overlap cost, with an analysis of what it keeps that point readouts lose. It outperforms the feature, coordinate keypoint and peak costs on block lifting and on average over three benchmarks.
\end{itemize}

%% file: sec/2_related.tex
\section{Related Work}
\label{sec:related}
\paragraph{Planning with frozen visual world models.}
The world models of DINO-WM~\cite{dinowm} and DDP-WM~\cite{ddpwm} are action-conditioned predictors trained on frozen DINOv2 patch features~\cite{dinov2}.
V-JEPA 2-AC~\cite{vjepa2} trains such a predictor on frozen V-JEPA 2 video features.
These models plan with CEM and score a candidate by the distance between its predicted features and those of the goal image.
LEAP~\cite{leap} also keeps the world model frozen and trains a decoder from its latent states to the state of the benchmark environment.
It adds the error of the decoded state to the latent distance and needs simulator states as labels.
Our readout predicts the distribution of any image point, is trained on rollouts of the model itself and needs no simulator state.

\paragraph{Uncertainty in model-based planning.}
PILCO~\cite{deisenroth2011pilco} learns a Gaussian process model of low-dimensional dynamics and propagates its uncertainty as Gaussian state distributions.
It minimizes the expected value of a saturating cost under these distributions.
PETS~\cite{chua2018pets} propagates particles through an ensemble of probabilistic networks and ranks CEM candidates by the average reward of the particles.
Both methods build the uncertainty into the dynamics model during training.
We read the uncertainty from the predictions of a frozen deterministic world model instead.
Visual Foresight~\cite{finn2017deep,ebert2017self,ebert2018visual} plans from images with a video prediction model.
The predicted pixel motion moves a probability distribution over the location of each pixel that the user designates.
Finn and Levine~\cite{finn2017deep} score a candidate by the probability that each designated pixel reaches its goal pixel.
Ebert \etal~\cite{ebert2017self} find this probability close to zero for long tasks and use the expected distance to the goal instead.
In these costs, the distributions come from pixel motion, and the user gives the designated and goal pixels.
We read the distribution of any query point from the predictions of a feature-space world model and the goal heatmap from the goal image.
We rank candidates by the probability that the prediction falls near the goal.
Our kernel keeps this probability informative far from the goal.

\paragraph{Keypoints for manipulation and planning.}
kPAM~\cite{kpam} and ReKep~\cite{rekep} specify manipulation tasks as costs and constraints on keypoints.
They move the keypoints with rigid transforms of the grasped objects.
KitF~\cite{kitf} learns dynamics on tracked keypoints and plans with model predictive control.
PointWorld~\cite{pointworld} learns a world model of the 3D motion of scene points from RGB-D observations.
It predicts a variance for each point during training~\cite{kendall2017uncertainties}, but its planning cost uses only the predicted positions.
DexWM~\cite{dexwm} supervises twelve hand keypoints with heatmaps when training its world model.
It adds the distance between the keypoint locations given by the predicted and goal heatmaps to its planning cost.
In these methods, the planning cost uses a single location for each point.
Our readout is added to a frozen world model, takes any query point and keeps the entire heatmap for the cost.

%% file: sec/3_method.tex
\section{Method}
\label{sec:method}

\begin{figure*}[t]
\centering
\includegraphics[width=\textwidth]{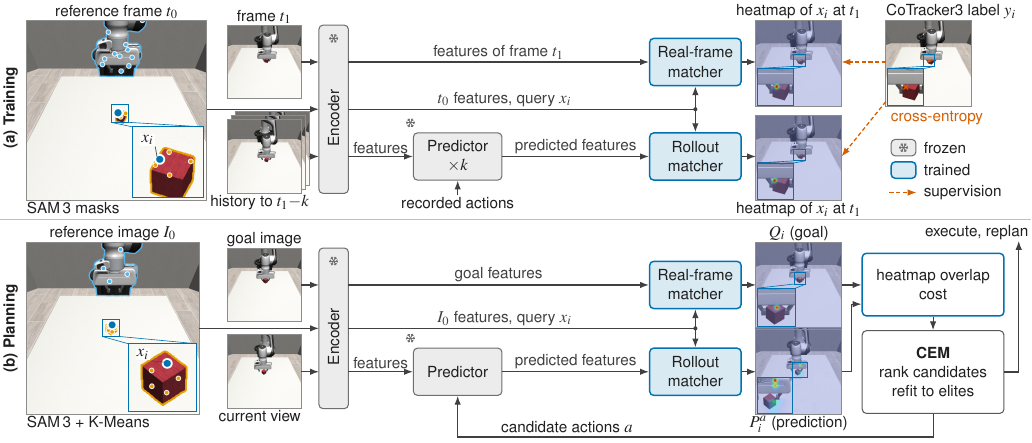}
\caption{\textbf{Probabilistic keypoint interface.} (a) CoTracker3 tracks points sampled in the SAM~3 masks of a reference frame and labels where each query point lies in a later frame. The real-frame matcher reads the real later frame. The rollout matcher reads the features that the frozen predictor rolls out under the recorded actions. (b) At planning time, the real-frame matcher reads the goal heatmap $Q_i$ once, and the rollout matcher reads $P^a_i$ for every candidate $a$. CEM ranks the candidates by the heatmap overlap cost. Only the front view and one query point are shown.}
\label{fig:pipeline}
\end{figure*}

The frozen world model encodes an observation into patch features and predicts the next features from past features and actions.
The cost must judge which candidate brings the task-relevant parts to the goal, but the predictions are only patch features.
We add readout networks that turn these predictions into heatmaps for the planning cost (\cref{fig:pipeline}).

\subsection{Problem and Evaluation Objective}
\label{sec:objective}

We place $K$ query points on the task-relevant objects in the reference image, which is taken at the start of the task.
The points of all views are numbered together.
Let $X_i$ be the image location of the $i$-th query point at the last step after executing a candidate $a$.
The same prediction can correspond to different real outcomes.
Given the prediction, $X=(X_1,\dots,X_K)$ is therefore a random vector, with joint distribution $\pi^a$ and $i$-th marginal $\pi_i^a$.
The goal heatmap $Q_i$ is the distribution of the location of the $i$-th point over the cells of the goal image.
It is read from the goal image by the real-frame matcher of \cref{sec:interface} and may have several peaks.
We measure how close two locations are with the Cauchy kernel $\kappa_\sigma(d)=(1+\|d\|^2/\sigma^2)^{-1}$ of scale $\sigma$.
The goal compatibility of the $i$-th point at location $x$ is
\begin{align}
u_i(x)&=\sum_y Q_i(y)\,\kappa_\sigma(x-y),\nonumber\\
f_i(x)&=\frac{\epsilon+u_i(x)}{\epsilon+u_i^*},
\label{eq:compat}
\end{align}
where $u_i^*=\max_x u_i(x)$ and $\epsilon$ is a small positive constant.
The compatibility $f_i$ equals 1 at the location that best agrees with the goal.
We call $\rho_i=-\log f_i$ the location loss.
A task with several parts needs all of them in place at the same time.
We give the $i$-th point a weight $w_i>0$ and let $\alpha_i=w_i/\sum_j w_j$.
The weighted geometric mean of the compatibilities measures whether all parts are in place together.
The joint goal cost is
\begin{equation}
R(a)=-\log\mathbb{E}_{X\sim\pi^a}\Bigl[\prod_i f_i(X_i)^{\alpha_i}\Bigr].
\label{eq:joint}
\end{equation}
Any part far from its goal pulls down the geometric mean, so $R$ is small only when all parts are likely in place together.
When each $f_i$ is the indicator of a goal region, $e^{-R}$ is the probability that all parts lie in their regions.
The kernel keeps $R$ finite and discriminative far from the goal.
Computing $R$ needs the joint distribution of the true locations, but we read only each point's marginal (\cref{sec:interface}).
\Cref{sec:information} shows that the marginals bound $R$ from both sides.

\subsection{Probabilistic Keypoint Interface}
\label{sec:interface}

The interface reads the location distribution of any query point from the predicted patch features.

\paragraph{Correspondence network.}
The network takes two frames and a query point, following COTR~\cite{cotr}.
A Transformer encoder jointly encodes the features of the reference image and of another time step.
A decoder reads this encoding through cross-attention at the continuous coordinate of each query point.
Query points do not attend to one another, so any pixel can serve as a query point.
With a coordinate head~\cite{toshev2014deeppose} trained by an L1 loss, the network is the keypoint matcher of the coordinate keypoint cost.
With a heatmap head~\cite{tompson2014joint,newell2016stacked}, it gives the probability of each image patch and of each cell within the patch.
Their product is a heatmap over the whole image, trained with the cross-entropy at the true location.

\paragraph{Keypoints for training.}
On Lift, we segment the task-relevant objects in the training videos with SAM~3~\cite{sam3} and sample points at random within the masks in proportion to mask area.
CoTracker3~\cite{cotracker3} tracks them in the RGB video.
Each training sample is a pair of frames at a random gap, and no simulator state is needed.
Other tasks use color segmentation.
At planning time, the query points are K-Means centers or random points in the object masks of the reference image.
On Lift, the only task-specific input is the text prompt that names the task objects for SAM~3.
The real-frame matcher is a heatmap readout trained on real frame pairs and gives the goal heatmap $Q_i$ from the goal image.

\paragraph{Rollout training.}
The real-frame matcher reads features of real images, but at planning time the readout must read predictions.
We therefore train a second heatmap readout, the rollout matcher, on rollouts of the world model itself~\cite{venkatraman2015improving,bengio2015scheduled}.
From a segment of real history in an offline trajectory, the frozen world model rolls out a random number $k$ of steps under the recorded actions.
The matcher reads the internal representations of the predictor at the last step, and the label is the tracked location in the real frame at step $k$.
Under the training actions, the cross-entropy is minimized by the conditional distribution $\pi_i^a$ of the true location given the prediction~\cite{gneiting2007strictly}.
At planning time, the rollout matcher gives the predicted heatmap $P_i^a$ of each candidate.
Its log-probabilities are divided by a temperature $T$~\cite{guo2017calibration}, fitted by negative log-likelihood on validation trajectories, and renormalized.

\subsection{Heatmap Overlap Cost}
\label{sec:overlap}

If each point takes its own expectation under its predicted heatmap $P_i^a$ in place of the joint expectation in \cref{eq:joint}, the cost needs only the marginals,
\begin{equation}
C(a)=-\sum_i\alpha_i\log\mathbb{E}_{P_i^a}f_i(X_i).
\label{eq:marginal}
\end{equation}
We implement it as the heatmap overlap cost, which compares the predicted and goal heatmaps of each query point,
\begin{align}
\mathcal{C}_{\mathrm{hm}}(a)&=-\sum_i w_i\log\bigl(\epsilon+o_i^a\bigr),\label{eq:overlap}\\
o_i^a&=\sum_{x,y}P_i^a(x)\,Q_i(y)\,\kappa_\sigma(x-y)=\sum_x P_i^a(x)\,u_i(x).\nonumber
\end{align}
The overlap $o_i^a$ is the expected kernel value between one location drawn from each of the two heatmaps~\cite{smola2007hilbert}.
With $W=\sum_i w_i$, we have $\alpha_i=w_i/W$. By \cref{eq:compat}, $\mathbb{E}_{P_i^a}f_i=(\epsilon+o_i^a)/(\epsilon+u_i^*)$, and hence $\mathcal{C}_{\mathrm{hm}}=W\,C-\sum_i w_i\log(\epsilon+u_i^*)$.
The second term does not depend on the candidate, so the two rank candidates identically.
Summing the negative logarithms of the overlaps of all points corresponds to the geometric mean of their compatibilities.

Three other costs use the same predicted heatmap.
The peak cost compares only the peak locations $m(\cdot)$ of the heatmaps, similar to the keypoint cost of DexWM~\cite{dexwm}.
The median cost compares their coordinate-wise medians instead.
The expected squared distance cost takes the expected squared distance from the predicted heatmap to the goal peak, like the expected distance of Ebert \etal~\cite{ebert2017self},
\begin{align}
\mathcal{C}_{\mathrm{peak}}(a)&=\sum_i w_i\bigl\|m(P_i^a)-m(Q_i)\bigr\|^2,\nonumber\\
\mathcal{C}_{\mathrm{exp}}(a)&=\sum_i w_i\sum_x P_i^a(x)\bigl\|x-m(Q_i)\bigr\|^2.
\label{eq:variants}
\end{align}

\subsection{Planning with CEM}
\label{sec:cem}

We compare with two more costs.
The feature cost is the most common cost for such world models~\cite{dinowm,ddpwm}.
It is the mean squared error between the predicted features $\hat z^a$ at the last step and the features $z_g$ of the goal image.
It includes the proprioceptive part when the state of the world model has one.
The coordinate keypoint cost also uses our interface.
It compares the coordinate $\hat p_i^a$ read by the keypoint matcher with the goal coordinate $\hat g_i$,
\begin{align}
\mathcal{C}_{\mathrm{feat}}(a)&=\MSE\bigl(\hat z^a,z_g\bigr),\nonumber\\
\mathcal{C}_{\mathrm{kp}}(a)&=\sum_i w_i\bigl\|\hat p_i^a-\hat g_i\bigr\|^2.
\label{eq:baselines}
\end{align}

All costs are optimized with the cross-entropy method commonly used with such world models~\cite{rubinstein1997cem,dinowm}.
Each iteration samples $N$ candidate action sequences from a diagonal Gaussian and refits the Gaussian to the $N_e$ elites of lowest cost.
After $I$ iterations, CEM executes an action sequence and replans.
On the benchmarks, it executes the mean of the elites, as the original planners of DINO-WM and DDP-WM do.
On Lift, it executes the lowest-cost candidate of the last iteration.

\section{What a Point Readout Discards}
\label{sec:information}

CEM acts only through the candidates that the cost selects.
We ask what the ranking of candidates loses when each predicted heatmap is reduced to a single location.
We consider separated outcomes, the goal neighborhood and several parts in turn, with proofs in Appendix~A of the supplementary material.
After each result, we state what it means for planning and which experiment tests it.

\input{figs/lift_reliability}

\subsection{Separated Outcomes}
\label{sec:info-gap}
A point readout tells where a part most likely ends up but not how likely it is to end up there.
Consider a query point that is lifted to $x_L$ with probability $q$ or stays on the table at a distant $x_T$, with heatmap $P_q=q\delta_{x_L}+(1-q)\delta_{x_T}$ and $f(x_L)>f(x_T)$.

\begin{proposition}
\label{prop:gap}
(i) For every $q>1/2$, the L1-optimal coordinate and the peak of $P_q$ are both $x_L$. The coordinate keypoint and peak costs therefore take one value for all such candidates and cannot rank them by $q$.
(ii) $C=-\log\bigl(qf(x_L)+(1-q)f(x_T)\bigr)$ strictly decreases in $q$, so the heatmap overlap cost ranks these candidates by $q$.
\end{proposition}

A point cost gives the same value to a grasp that may fail and to a secure grasp.
CEM then chooses between them arbitrarily.
A more accurate coordinate regressor does not remove this tie, because the probability of lifting is not part of its output.
The tie of (i) appears on real predictions.
On 5,117 Lift candidates executed in the simulator (\cref{sec:exp-info}), the average lift-region probability of the heatmaps follows the observed lift rate.
The fraction of coordinate readouts in the lift region takes almost only the values 0, 0.5 and 1 (\cref{fig:lift-prob}a).
On the same predicted heatmaps, the heatmap overlap cost succeeds in 86.3\% of the two-stage Lift episodes (\cref{sec:exp-setup}), compared with 73.5\% for the peak cost ($p=0.016$) and 82.4\% for the median cost (\cref{tab:ablation,tab:paired}).

\subsection{Near the Goal}
\label{sec:info-local}
Intuitively, when two candidates place a part at the same location near the goal, the more certain candidate is more likely to have the part truly at the goal.
Let $x_i^*$ be a nondegenerate maximizer of $u_i$ and $H_i=\nabla^2\rho_i(x_i^*)$.
Consider candidates whose centers and spreads both lie near the goal at a scale $s$, with $X_i=x_i^*+sZ_i$.
The joint distribution of $Z=(Z_1,\dots,Z_K)$ is fixed and bounded, and its components may be arbitrarily correlated.
A point readout that scores the mean location of each point with the same compatibility has the cost $C_{\mathrm{pt}}=\sum_i\alpha_i\rho_i(\mathbb{E}X_i)$.

\begin{proposition}
\label{thm:local}
Let $\Sigma_i$ be the covariance of $X_i$. As $s\to0$,
\begin{align}
C-C_{\mathrm{pt}}&=\tfrac12\sum_i\alpha_i\operatorname{tr}\bigl(H_i\Sigma_i\bigr)+O(s^3),\nonumber\\
R-C&=O(s^4).
\label{eq:local}
\end{align}
\end{proposition}

For a point goal, the second-order term is approximately the location variance divided by $\sigma^2$.
For small $s$, the candidate with the larger dispersion term therefore has the higher cost when the mean locations are equal.
Any point readout that places both candidates at the same locations gives them the same cost (Corollary~2 in Appendix~A.3).
For symmetric unimodal heatmaps, the mean, the median and the mode coincide, so this holds for all three readouts.
Near the goal, a point readout cannot prefer the candidate whose parts are more certainly in place.
The heatmap overlap cost charges this dispersion through the curvature of the location loss.

The expected squared distance cost also keeps this term, since $\mathbb{E}\|X-g\|^2=\|\mathbb{E}X-g\|^2+\operatorname{tr}\operatorname{Cov}X$ for the goal peak $g$.
But it sees a heatmap only through its mean and covariance.
Probability at a distance $D$ from the goal raises it in proportion to $D^2$.
Mixing in a fraction $\eta$ of any distribution raises $C$ by at most $-\log(1-\eta)$ (Appendix~A.8).
On the same heatmaps, the expected squared distance cost succeeds in 76.5\% of the two-stage and 24.5\% of the low-goal Lift episodes.
The heatmap overlap cost reaches 86.3\% and 69.6\% and wins 50 low-goal episodes to 4 (\cref{tab:ablation,tab:paired}).
Keeping the dispersion is therefore not enough.
The heatmap overlap cost also keeps how the probability is split between distinct outcomes.
Its Cauchy kernel keeps distant candidates apart.
A Gaussian kernel of the same width falls below $\epsilon$ beyond about $5\sigma$ from the goal and ranks successful candidates ahead of failed ones less often at 13 of 17 starts.

\subsection{Marginal Heatmaps and the Joint Goal}
\label{sec:info-joint}
The joint goal asks for all parts to be in place in the same outcome.
The heatmaps only say how likely each part is to be in place.
The gap between the two is small for concentrated heatmaps and grows as they spread out.
Let $\Delta(a)=\sum_i\alpha_i\bigl(\mathbb{E}_{P_i^a}\rho_i(X_i)+\log\mathbb{E}_{P_i^a}f_i(X_i)\bigr)$ be the weighted sum of the Jensen gaps of the points.
It is nonnegative and computed from the heatmaps.

\begin{proposition}
\label{thm:sandwich}
If $P_i^a=\pi_i^a$, then for any joint distribution $\pi^a$ consistent with these marginals,
\begin{equation}
C(a)\le R(a)\le C(a)+\Delta(a).
\label{eq:sandwich}
\end{equation}
\end{proposition}

\begin{corollary}
\label{cor:select}
Let $\mathcal{E}$ be the $N_e$ candidates with the lowest $C$ in a batch and $r_{(N_e)}$ the $N_e$-th smallest value of $R$ in the batch.
Then every candidate $a\in\mathcal{E}$ satisfies $R(a)\le r_{(N_e)}+\Delta(a)$.
If $C(a)+\Delta(a)<C(b)$, then $R(a)<R(b)$.
\end{corollary}

\Cref{thm:sandwich} does not assume independent points or Gaussian or continuous distributions.
It holds exactly on the discrete grid of the heatmaps, and $R=C$ for a single query point.
With the joint goal utility $V=e^{-R}$, the candidate $\hat a$ of lowest $C$ satisfies $V(\hat a)\ge e^{-\Delta(\hat a)}\max_a V(a)$.
Near the goal, \cref{thm:local} gives $R-C=O(s^4)$, two orders below the dispersion term that a point readout misses.
The heatmap overlap cost therefore needs no model of how the parts move together.
For each candidate, $\Delta$ shows whether leaving out the joint distribution can change its rank.
On the executed Lift candidates, the condition of \cref{cor:select} fixes the order of 96\% of the candidate pairs under any correlation across points.
The candidate of lowest $C$ has a median $\Delta$ of 0.015.
Reducing each heatmap to a point loses more than ignoring this correlation.
Within 10 pixels of the goal peak, the dispersion term missed by the point readout is 4 to 8 times $\Delta$ in median.
It exceeds $\Delta$ at every start (Appendix~D.5).

\paragraph{From pairs to elites.}
These results order pairs of candidates, but CEM keeps a set of elites.
Appendix~A.13 bounds how much less likely the elites selected by $C$ are to succeed than the most reliable candidates.
The bound depends on how well the heatmaps are calibrated and how closely $C$ follows their outcome probabilities.
A calibrated readout therefore lets CEM keep reliable candidates even when it does not sample them more often.
In the first rounds of 34 two-stage episodes with 2 seeds per demonstration, CEM samples graspable candidates at similar rates under both keypoint costs, 58\% and 59\% at iteration 10 (\cref{fig:lift-prob}b and Appendix~D.7).
However, the lowest-cost candidate is graspable in 26 rounds under the heatmap overlap cost and in 19 under the coordinate keypoint cost ($p=0.039$, sign test).

%% file: figs/lift_reliability.tex
\begin{figure}[t]
\centering
\includegraphics[width=\linewidth]{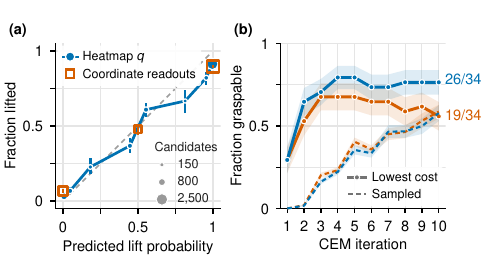}
\caption{\textbf{The heatmaps track lifting, and CEM selects graspable candidates more often with the heatmap overlap cost.}
(a) The 5,117 executed Lift candidates, binned by the average lift-region probability $q$ of the 10 block points.
Dots give the lift rate with 95\% Wilson intervals and are sized by the number of candidates.
Squares give the coordinate readouts, and the dashed line marks perfect calibration.
(b) The first rounds of the 34 two-stage episodes, with the heatmap overlap cost in blue and the coordinate keypoint cost in orange.
Solid lines give the share of rounds whose lowest-cost candidate is graspable, dashed lines the share of graspable sampled candidates, and shading the standard error.}
\label{fig:lift-prob}
\end{figure}

%% file: sec/4_experiments.tex
\section{Experiments}
\label{sec:exp}

\begin{table}[t]
\centering\footnotesize
\setlength{\tabcolsep}{3pt}
\caption{\textbf{Compared costs.} All keypoint costs use the same query points and goal images. The peak, median, expected squared distance and heatmap overlap costs score the same predicted heatmaps, so their differences come only from the scoring.}
\label{tab:costs}
\begin{tabular}{@{}lll@{}}
\toprule
Cost & Readout network & Scoring \\
\midrule
Feature & none & Feature MSE \\
Coordinate keypoint & keypoint matcher & Sq.\ coordinate distance \\
Peak & rollout matcher & Sq.\ peak distance \\
Median & rollout matcher & Sq.\ median distance \\
Exp.\ sq.\ distance & rollout matcher & Exp.\ sq.\ dist.\ to goal peak \\
Heatmap overlap & rollout matcher & Overlap of entire heatmaps \\
\bottomrule
\end{tabular}
\end{table}

\begin{table*}[t]
\centering\footnotesize
\caption{\textbf{The heatmap overlap cost has the highest or tied-highest final success rate in every setting and the highest average.} On Lift, the success rate (\%) is at the end over 102 episodes, and elsewhere after the first round / at the end over 150 goals.}
\label{tab:main}
\begin{tabular}{@{}llcccc@{}}
\toprule
Task & World model & Feature $\uparrow$ & Coordinate keypoint $\uparrow$ & Peak $\uparrow$ & Heatmap overlap $\uparrow$ \\
\midrule
Lift, low goal             & DINO-WM & 16.7 & 58.8 & 55.9 & \textbf{69.6} \\
Lift, two-stage goal       & DINO-WM & -- & 59.8 & 73.5 & \textbf{86.3} \\
\midrule
\multirow{2}{*}{PushT}     & DINO-WM & 82.7 / 86.0 & 86.0 / 89.3 & 84.7 / 87.3 & \textbf{88.7} / \textbf{94.0} \\
                           & DDP-WM  & 75.3 / 80.0 & 80.0 / 85.3 & 82.7 / 86.7 & \textbf{87.3} / \textbf{92.7} \\
\multirow{2}{*}{Wall}      & DINO-WM & 56.0 / 93.3 & 76.7 / 92.7 & 86.0 / 92.7 & \textbf{87.3} / \textbf{94.0} \\
                           & DDP-WM  & 58.0 / 88.0 & 66.7 / 90.7 & 80.0 / 92.7 & \textbf{85.3} / \textbf{94.0} \\
\multirow{2}{*}{PointMaze} & DINO-WM & 80.0 / 94.7 & 91.3 / 97.3 & \textbf{93.3} / 98.7 & 92.7 / \textbf{99.3} \\
                           & DDP-WM  & 80.0 / 92.7 & 79.3 / 96.7 & 86.7 / \textbf{97.3} & \textbf{88.7} / \textbf{97.3} \\
\midrule
\multicolumn{2}{@{}l}{Average of the last six rows} & 72.0 / 89.1 & 80.0 / 92.0 & 85.6 / 92.6 & \textbf{88.3} / \textbf{95.2} \\
\bottomrule
\end{tabular}
\end{table*}

\subsection{Setup}
\label{sec:exp-setup}

\paragraph{Tasks and world models.}
We evaluate on two manipulation tasks, lifting a block (Lift) and the contact-rich PushT, and on two navigation tasks, Wall and PointMaze.
Lift comes from robosuite~\cite{robosuite,todorov2012mujoco} with a front and a side camera.
PushT~\cite{florence2021implicit,chi2023diffusion}, Wall and PointMaze~\cite{fu2020d4rl} come from the benchmark of DINO-WM~\cite{dinowm} and use the world models released by the authors of DINO-WM and DDP-WM~\cite{ddpwm} with their original CEM settings.
On Lift, a DINO-WM world model is trained on the training demonstrations of robomimic~\cite{mandlekar2021robomimic} with the original procedure.
It is frozen before the readouts are trained.

\paragraph{Goals and evaluation.}
PushT, Wall and PointMaze follow the original evaluation procedure with three seeds of 50 goals each.
Lift uses the 17 held-out demonstrations whose block can be lifted with 4-dimensional actions.
Each is planned with 6 seeds from its initial state, giving 102 episodes.
An episode succeeds when the block center is more than 4\,cm above the table, about 2\,cm above its resting height.
We report success after the first planned action sequence, which reflects the ranking most directly, and at the end.
No test demonstration is used for training (Appendix~B.6).

\paragraph{Two goal settings on Lift.}
In the two-stage setting, the goal image is the last frame of the demonstration.
The first round brings the arm to the block, and the second round lifts it.
Both rounds derive their goals automatically from the goal image and the current frame.
All keypoint costs use this construction, each with its own readout (Appendix~B.4).
In the low-goal setting, the goal image is the first frame in which the block center is 6\,cm above the table.
One cost is used throughout.
The feature cost succeeds in no Lift episode when its goal image is the last frame of the demonstration.
We therefore compare it with the other costs in the low-goal setting.

\begin{table}[t]
\centering\footnotesize
\setlength{\tabcolsep}{3pt}
\caption{\textbf{Paired comparisons of the heatmap overlap cost with each other cost.} Overlap and Other give the success rates (\%) of the two costs on the same episodes or goals, and Diff.\ is their difference in points. Tests pair the 17 Lift demonstrations and the benchmark (Bench.) goals within each task and world model. The 3 Lift groups are both goal settings and the two-stage setting with 100 candidates. Feature, objects uses only object patches, and No rollout tr.\ reads predictions with the real-frame matcher.}
\label{tab:paired}
\begin{tabular}{@{}llcccc@{}}
\toprule
Other cost & Setting & Overlap & Other & Diff. & $p$ \\
\midrule
Coordinate & Lift, two-stage & 86.3 & 59.8 & $+26.5$ & 0.019 \\
Coordinate & Bench., round 1 & 88.3 & 80.0 & $+8.3$ & $<10^{-4}$ \\
Peak & Lift, 3 groups & 76.8 & 63.1 & $+13.7$ & 0.002 \\
Peak & Bench., round 1 & 88.3 & 85.6 & $+2.8$ & 0.011 \\
Median & Lift, 2 settings & 77.9 & 67.6 & $+10.3$ & 0.030 \\
Exp.\ sq.\ dist. & Lift, low goal & 69.6 & 24.5 & $+45.1$ & $1.2{\times}10^{-4}$ \\
Feature, objects & Lift, low goal & 69.6 & 7.8 & $+61.8$ & $3{\times}10^{-5}$ \\
No rollout tr. & Lift, low goal & 69.6 & 44.1 & $+25.5$ & 0.028 \\
\bottomrule
\end{tabular}
\end{table}

\begin{table}[t]
\centering\small
\setlength{\tabcolsep}{3.2pt}
\caption{\textbf{With a reduced planning budget or a more distant goal, the heatmap overlap cost still has the highest success rate.} Final success rate (\%) with goals 25 or 50 environment steps away. The 1/18 budget uses 50 candidates, 5 elites and 10 iterations.}
\label{tab:budget}
\footnotesize
\begin{tabular}{@{}llccc@{}}
\toprule
Task, world model & Budget & Feature & Coordinate & Overlap \\
\midrule
PushT 25, DINO-WM & full & 86.0 & 89.3 & \textbf{94.0} \\
                  & 1/18 & 73.3 & 82.0 & \textbf{90.0} \\
PushT 25, DDP-WM  & full & 80.0 & 85.3 & \textbf{92.7} \\
                  & 1/18 & 64.0 & 75.3 & \textbf{88.0} \\
PushT 50, DINO-WM & full & 26.7 & 50.0 & \textbf{54.7} \\
                  & 1/18 & 15.3 & 35.3 & \textbf{41.3} \\
PushT 50, DDP-WM  & full & 15.3 & 42.0 & \textbf{52.7} \\
                  & 1/18 & 7.3 & 36.0 & \textbf{39.3} \\
\bottomrule
\end{tabular}
\end{table}

\paragraph{Planning.}
CEM samples 300 candidates and keeps 30 elites, with 30 iterations on PushT and 10 elsewhere.
The benchmark tasks plan and execute 5 model steps per round for at most 3 rounds.
Lift is planned in two rounds with 5 block points and 4 arm points per view.

\paragraph{Compared costs.}
\Cref{tab:costs} lists the compared costs.
The feature cost uses no query points, so comparing with it measures what the interface and the cost bring together.

\subsection{Planning Success}
\label{sec:exp-main}

With the same world model and planner, the heatmap overlap cost has the highest success rate on Lift in both goal settings (\cref{tab:main}).
On PushT, Wall and PointMaze, it has the highest or tied-highest final success rate in all six combinations of task and world model.
It also has the highest average after the first round and at the end.
\Cref{tab:paired} pairs the heatmap overlap cost with each other cost on the same episodes or goals.
On the benchmarks, it also beats the coordinate keypoint cost by 3.2 points at the end.
It beats the feature cost by 16.3 points after the first round and by 6.1 at the end, all with $p<10^{-4}$.
The gap to the feature cost does not come from the object masks.
On the patches of the task objects alone, the feature cost succeeds in 7.8\% of the low-goal episodes, lower than 16.7\% on the whole image (Appendix~D.3).
The heatmap overlap cost differs from the coordinate keypoint cost in the readout training and in the scoring, and \cref{tab:ablation} changes one with the other fixed.
Scoring the entire heatmap instead of its peak gives the larger gain in the two-stage setting, and rollout training in the low-goal setting.
On the benchmarks, the rollout-matcher peak cost already beats the coordinate keypoint cost after the first round.
The heatmap overlap cost improves on it further.

With 1/18 of the planning budget, the heatmap overlap cost reaches success rates of 90.0\% and 88.0\% on PushT with goals 25 steps away (\cref{tab:budget}).
These exceed the 89.3\% and 85.3\% of the coordinate keypoint cost at full budget.
At the same budget, the peak cost reaches 83.3\% and 76.0\%.
A readout takes about 0.45\,ms per candidate, and a CEM iteration on DINO-WM takes 1.03 to 1.04 times as long as with the feature cost.
With goals 50 steps away and at most 6 rounds, the heatmap overlap cost has the highest success rate, 54.7\% and 52.7\% at full budget.

\subsection{Outcome Probabilities in the Heatmaps}
\label{sec:exp-info}

\Cref{prop:gap} needs heatmaps that give the probability of each outcome, which we test on Lift.
We execute 5,117 action sequences in the simulator, starting one step before grasping in each test demonstration.
The average lift-region probability $q$ is the probability that the predicted heatmap of a point assigns to its lift region, averaged over the 10 block points of the two views.
The lift region of a point is the half-plane of the image beyond the projection of the point raised by 2\,cm, the height of the success criterion (Appendix~D.6).
The lift rate rises monotonically with $q$, from 3\% for $q<0.02$ to 91\% for $q\ge0.98$ (\cref{fig:lift-prob}a).
The probability $q$ predicts lifting with a median AUC of 0.95 over starts, compared with 0.92 for the coordinate readouts and for the peaks, and is higher than both at every start.

As a control, we replace the world model of PushT by the exact simulator.
It replays the executed actions and the candidate from the initial state and renders the last frame.
The feature cost encodes this frame with DINO-WM, and the two keypoint costs read its features with the matchers trained on real frames (Appendix~H).
Even with the exact simulator, the feature cost remains the lowest of the three at 86.7\%.
It solves 1 goal that the heatmap overlap cost misses, against 10 the other way ($p=0.012$).
Against the coordinate keypoint cost, the counts are 2 and 13 ($p=0.007$).
The gap to the feature cost therefore does not come from errors of the world model.
The coordinate keypoint and heatmap overlap costs reach 94.0\% and 92.7\% on the simulator and 89.3\% and 94.0\% on DINO-WM.

\subsection{Scoring and Selection}
\label{sec:exp-same}

\begin{table}[t]
\centering\footnotesize
\setlength{\tabcolsep}{4pt}
\caption{\textbf{Scoring and rollout training on Lift.} Lift success rate (\%) over 102 episodes. The row without rollout training reads the predictions with the real-frame matcher, and the others the rollout matcher. With 100 candidates in the two-stage setting, the heatmap overlap and peak costs reach 74.5\% and 59.8\%.}
\label{tab:ablation}
\begin{tabular}{@{}lcc@{}}
\toprule
Scoring & Two-stage & Low goal \\
\midrule
Peak & 73.5 & 55.9 \\
Median & 82.4 & 52.9 \\
Exp.\ sq.\ distance & 76.5 & 24.5 \\
Overlap, no rollout training & 83.3 & 44.1 \\
Heatmap overlap & \textbf{86.3} & \textbf{69.6} \\
\bottomrule
\end{tabular}
\end{table}

\Cref{tab:ablation} tests whether the scoring alone makes CEM select more reliable candidates.
The heatmap overlap cost has the highest success rate in every column.
The median cost isolates the effect of the scoring.
The rollout matcher estimates the conditional distribution of the true location, so the coordinate-wise median of its heatmap approximates the output of an ideal L1 coordinate head trained on the same rollouts.
The median and heatmap overlap costs therefore share the readout and its training data and differ only in how they score.
The keypoint matcher is trained on real frames instead.
\Cref{sec:information} relates these gaps to the analysis.

\subsection{Training and Calibration of the Readout}
\label{sec:exp-readout}

Without rollout training, the predicted heatmaps are read by the real-frame matcher.
Rollout training raises the success rate in both Lift settings (\cref{tab:ablation}) and from 63\% to 92\% on 100 goals of Wall with DDP-WM.
On the first-round CEM candidates, the front-view 50\% and 90\% regions contain the true location in 0.50 and 0.89 to 0.92 of the cases in every iteration.
For the side view, the fractions are 0.45 to 0.49 and 0.83 to 0.89.
The front-view negative log-likelihood is 1.45, below the 1.75 of the real-frame matcher.
With $\sigma$ of 8 or 32 pixels instead of 16, the low-goal success rate does not change significantly ($p\ge0.17$).
The ranking stays the same for $\epsilon$ from $10^{-8}$ to $10^{-3}$ and keeps a rank correlation of at least 0.989 for 3 to 8 points per view (Appendix~F.5).

%% file: sec/5_conclusion.tex
\section{Scope and Limitations}
\label{sec:limits}
All evaluations are in simulation, with Lift as a three-dimensional and PushT as a contact-rich manipulation task.
The method needs the names of the task objects for SAM~3 and readout networks trained on image tracks for each world model.
It needs no simulator state, so the same training applies to recorded robot video.
On the Lift demonstrations, the CoTracker3 labels deviate from the simulator ground truth by about 1 pixel in median on $1024\times1024$ images.
Calibration was measured on the first-round CEM candidates of Lift and on the executed candidates of \cref{sec:exp-info}.

\section{Conclusion}
\label{sec:conclusion}
We studied which information a planning cost should keep from the predictions of a frozen world model.
Given these predictions, the probabilistic keypoint interface reads out how likely each outcome is.
The heatmap overlap cost ranks candidates by these probabilities.
Our analysis shows that point readouts discard two things this ranking needs, the split of probability between distant outcomes and the dispersion near the goal.
It also shows why scoring each point by its own heatmap suffices.
Each of these effects appears on real predictions.
With the same world model and planner, the heatmap overlap cost has the highest or tied-highest final success rate in every setting of \cref{tab:main,tab:budget}.